\documentclass[accepted]{melba}
\usepackage{mwe} % to get dummy images, only for the example
\usepackage{amsmath,amsfonts}
\melbaid{2025:033}  % This is provided upon by the publishing editor
\doi{10.59275/j.melba.2025-5242}
\melbaauthors{Akis Linardos et al.}  % Note: this one is also used to set the pdf 'authors' metadata
\email{spbakas@iu.edu}
\volume{3}
\firstpageno{757}  % Communicated by the publishing editor
\melbayear{2025}  % The publication year
\datesubmitted{2025-4-25}  % Date submitted to MELBA: mm/yyyy
\datepublished{2025-12-4}  % Today's date: mm/yyyy

\usepackage{multirow}
\usepackage{tikz}
\usepackage{subfig}

\definecolor{ReMIC}{rgb}{0.867, 0.518, 0.322}      % Deep Orange
\definecolor{Flair}{rgb}{0.333, 0.659, 0.408}        % Deep Green
\definecolor{HTTUAS-RL}{rgb}{0.506, 0.446, 0.702}    % Deep Purple 
\definecolor{HTTUAS-REC}{rgb}{0.867, 0.322, 0.322}   % Deep Red 
\definecolor{SNU}{rgb}{0.576, 0.471, 0.376}          % Deep Brown 
\definecolor{rigg}{rgb}{0.298, 0.447, 0.690}         % Deep Blue

\melbaspecialissue{}
\melbaspecialissueeditors{}

\ShortHeadings{MELBA Journal Sample Article}{Linardos et al}

\title{The MICCAI Federated Tumor Segmentation (FeTS) Challenge 2024:
Efficient and Robust Aggregation Methods for Federated Learning}

\author{
  \firstname Akis \surname Linardos\aff{1,2}\orcid{0000-0002-5470-1908},
  \firstname Sarthak \surname Pati\aff{1,2,3},
  \firstname Ujjwal \surname Baid\aff{1,2}\orcid{0000-0001-5246-2088},
  \firstname Brandon \surname Edwards\aff{4},
  \firstname Patrick \surname Foley\aff{4},
  \firstname Kevin \surname Ta\aff{4},
  \firstname Verena \surname Chung\aff{5},
  \firstname Micah \surname Sheller\aff{3,4},
  \firstname Muhammad Irfan \surname Khan\aff{6},
  \firstname Mojtaba \surname Jafaritadi\aff{7},
  \firstname Elina \surname Kontio\aff{6},
  \firstname Suleiman \surname Khan\aff{6},
  \firstname Leon \surname M\"achler\aff{8},
  \firstname Ivan \surname Ezhov\aff{9},
  \firstname Suprosanna \surname Shit\aff{9},
  \firstname Johannes C. \surname Paetzold\aff{10},
  \firstname Gustav \surname Grimberg\aff{11},
  \firstname Manuel A. \surname Nickel\aff{9},
  \firstname David \surname Naccache\aff{8},
  \firstname Vasilis \surname Siomos\aff{12},
  \firstname Jonathan \surname Passerat-Palmbach\aff{13},
  \firstname Giacomo \surname Tarroni\aff{12,13},
  \firstname Daewoon \surname Kim\aff{14},
  \firstname Leonard L. \surname Klausmann\aff{15},
  \firstname Prashant \surname Shah\aff{4},
  \firstname Bjoern \surname Menze\aff{16},
  \firstname Dimitrios \surname Makris\aff{17}\orcid{0000-0001-6033-1731},
  \firstname Spyridon \surname Bakas\aff{1,2,3,17,18,19}\orcid{0000-0001-8734-6482}
}

\affiliations{%
  \num 1 \addr Department of Pathology and Laboratory Medicine, Indiana University School of Medicine, Indianapolis, IN, USA \\
  \num 2 \addr Center for Federated Learning in Medicine, Indiana University School of Medicine, Indianapolis, IN, USA \\
  \num 3 \addr Medical AI Group, MLCommons, San Francisco, CA, USA \\
  \num 4 \addr Intel Corporation, Santa Clara, CA, USA \\
  \num 5 \addr Sage Bionetworks, Seattle, WA, USA \\
  \num 6 \addr Turku University of Applied Sciences, Turku, Finland \\
  \num 7 \addr Stanford University, Stanford, CA, USA \\
  \num 8 \addr Ecole Normale Supérieure, Paris, France \\
  \num 9 \addr Technical University of Munich, Munich, Germany \\
  \num 10 \addr Weill Cornell Medicine, New York, USA \\
  \num 11 \addr Ezri AI Labs, Paris, France \\
  \num 12 \addr City St George’s, University of London, UK \\
  \num 13 \addr Imperial College London, London, UK \\
  \num 14 \addr Seoul National University, Seoul, South Korea \\
  \num 15 \addr Ostbayerische Technische Hochschule (OTH) Regensburg, Germany \\
  \num 16 \addr Universität Zürich, Zürich, Switzerland \\
  \num 17 \addr Kingston University London, London, UK \\
  \num 18 \addr Departments of Radiology and Imaging Sciences; Neurological Surgery; Biostatistics and Health Data Science, Indiana University School of Medicine, Indianapolis, IN, USA \\
  \num 19 \addr Department of Computer Science, Luddy School of Informatics, Computing and Engineering, Indiana University, Indianapolis, IN, USA \\
}

\abstract{%   <- trailing '%' for backward compatibility of .sty file
	We present the design and results of the MICCAI Federated Tumor Segmentation (FeTS) Challenge 2024, focusing on federated learning (FL) for glioma sub-region segmentation in multi-parametric MRI scans, and evaluating novel weight aggregation methods for increased robustness and efficiency. Participating methods from six teams are evaluated using a standardized FL setup and a multi-institutional dataset derived from the BraTS glioma benchmark---a dataset consisting of 1,251 training cases, 219 validation cases, and 570 hidden test cases, with segmentations of enhancing tumor (ET), tumor core (TC), and whole tumor (WT). Teams are ranked by a cumulative scoring system accounting for segmentation performance---measured by Dice Similarity Coefficient (DSC) and 95th percentile Hausdorff Distance (HD95)---and communication efficiency assessed through the convergence score. A PID-controller-based approach emerges as the top-performing method, achieving a mean DSC of 0.733, 0.761, and 0.751 for ET, TC, and WT, respectively, with corresponding HD95 values of 33.922mm, 33.623mm, and 32.309mm, while also being the most efficient with a convergence score of 0.764. These results contribute to ongoing advances in FL, building on top-performers from previous challenge iterations and surpassing them, highlighting PID controllers as powerful mechanisms for stabilizing and optimizing weight aggregation in FL. The challenge code is available at~\url{https://github.com/FeTS-AI/Challenge}.}


\begin{document}

% top matter
\twocolumn[\maketitle]
% comment the preceedings and uncomment the following if the authors list + abstract is longer than one page
% \maketitle
% \twocolumn

% Introduction (or first section)
% \rule{\textwidth}{1pt}

\section{Background}

    While AI has been making strides in all fields, its applicability in healthcare has been mainly hindered by data scarcity, with most studies focusing on single-center data \citep{rajpurkar2022ai, kelly2019key}, not able to capture the diversity across patient populations. Models produced in such a restricted manner have questionable generalizability in real-world applications, where data significantly varies from one site to the next. To address the scarcity, collaborative studies are essential, and to remain respectful of privacy constraints (such as HIPPA and GDPR \citep{hipaa,gdpr}), a realizable way forward is through Federated Learning (FL): a framework that distributes models across sites, learns locally from institutional data, while alleviating the obvious privacy risks of data-sharing and hence acting as a catalyst for multi-site healthcare partnerships \citep{mcmahan2017communication,fl_1,fl_2,fl_3,sheller2020federated, yang2024federated, pati2024privacy}. This way models may capture information from the high diversity of the real-world data, remaining fair across different populations.

    It is important to note that while FL mitigates the obvious privacy risks by keeping data local, it does not guarantee privacy by itself \citep{pati2024privacy,zhao2025federation}. Model updates can still leak sensitive information through, for example, membership inference attacks~\citep{hu2022membership, zhang2022survey}. Complementary approaches to strengthen privacy in federated setups include secure aggregation~\citep{fereidooni2021safelearn,so2022lightsecagg,rathee2023elsa}, differential privacy~\citep{el2022differential,adnan2022federated}, homomorphic encryption~\citep{xie2024efficiency,aziz2023exploring}, and confidential computing\footnote{\url{https://confidentialcomputing.io/}}.
    
    For the work presented here, the use case is glioma segmentation, encompassing both low-grade and high-grade gliomas---including glioblastoma, the most common and aggressive type of adult brain tumor. Glioblastoma, despite multimodal treatments involving surgical resection, radiation, and chemotherapy, has a median survival of about 15 months, with less than 10\% of patients surviving for over 5 years \citep{cbtrus_2019}. The poor prognosis is largely a consequence of glioblastoma complexity, whose pathological heterogeneity leads to treatment resistance~\cite{bakas2024brats,villanueva2024artificial,bakas2024artificial}.
    Routine diagnosis and response assessment in glioblastoma patients is carried out through radiologic imaging (i.e., magnetic resonance imaging (MRI))~\citep{shukla2017advanced}, through which the tumor subregions may be delineated for follow-up computational analyses and personalized diagnostics \citep{pati2020reproducibility}. To enable robust tumor subregion delineation, the International Brain Tumor Segmentation (BraTS) benchmark/challenge \citep{menze2014multimodal,bakas2018identifying,bakas2017advancing,bakas2017segmentation_1,bakas2017segmentation_2, baid2021rsna} has been at the forefront of providing high-quality data and an end-to-end open-source framework that fosters a benchmark environment for fair algorithmic evaluation. It used clinically acquired, multi-parametric MRI (mpMRI), and the evaluated algorithms are publicly available for use by the scientific community \citep{bakas2015glistrboost,zeng2016segmentation,kamnitsas2017efficient,isensee2018nnu,mckinley2018ensembles}.
    
    The Federated Tumor Segmentation (FeTS) challenge 2021 leveraged the BraTS glioma dataset to present the first challenge ever proposed for FL. The FeTS challenge in 2021 focused on constructing and evaluating a consensus model for the segmentation of gliomas, while its continuation in FeTS 2022 built on the foundation laid by its predecessor, further refining the federated learning techniques and expanding the collaborative network for an even larger dataset \citep{zenk2025towards}. Since the first study of FL in healthcare \citep{sheller2019multi,sheller2020federated} and following the FeTS 2022 challenge, 
    % \textbf{REF: cite fets challenge paper when it's on arxiv},
    there have been numerous studies across medical imaging fields that focused on federated learning either in a simulation set up where the ``different sites'' are actually running on a single machine \citep{linardos2022federated, ro2021fedjax, li2022federated, adnan2022federated} or real-world application, where the federated learning set up is deployed across actual sites, bringing together cohorts that span the globe \citep{pati2022federated}. Along the trajectory of this growth, there have also been multiple notable tools that foster FL research, such as the MedPerf for federated benchmarking of AI models ‘in the wild' \citep{karargyris2023federated}, fedJAX for simulation-focused research \citep{ro2021fedjax}, and multiple libraries for FL development ~\citep{foley2022openfl,roth2022nvidia,beutel2020flower,ziller2021pysyft, pati2022federated_tool}.
    
    Previous FeTS Challenges (2021 and 2022)~\cite{zenk2025towards} have presented tasks in two scenarios: one task in an environment that replicates federated learning conditions within one machine using BraTS, and a second task where evaluation is carried out across a real-world federation. Building on the insights from these previous challenges, the FeTS Challenge 2024, the third iteration of this challenge, shifts its focus exclusively to ``optimal weight aggregation'', testing innovations on the federated aggregation algorithm on a single machine, without the hurdle of real-world deployment. The primary goal of this challenge is to further refine the FL approach by optimizing the aggregation of model weights from different institutions, thereby improving the overall performance and robustness of the consensus models. This focus aims to address some of the remaining challenges in FL, particularly how best to combine the learned parameters from diverse data sources without compromising data privacy.

    \begin{figure*}[ht!]
        \centering
        \includegraphics[width=\textwidth]{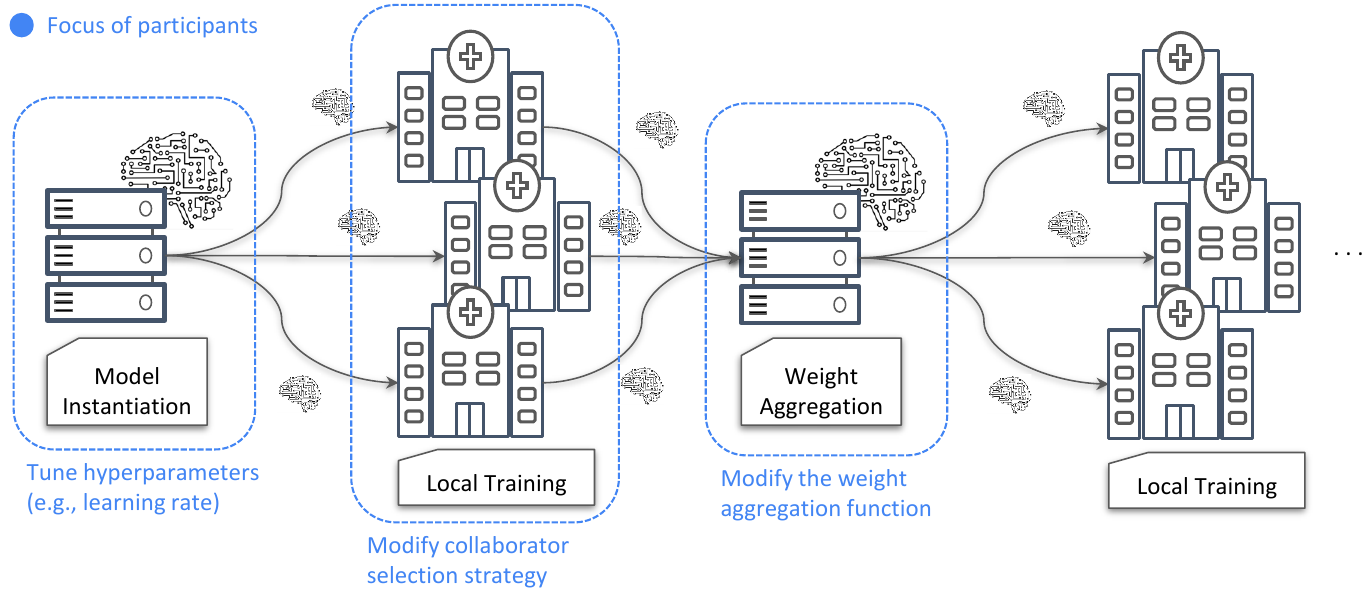}
        \caption{A schematic of the FL-based learning process. The FeTS Challenge tasks participants with contributing their innovations at the three levels: a. hyperparameter tuning, b. collaborator selection, c. weight aggregation function. Note that for this challenge, the individual sites illustrated here are actually data partitions within one computational environment rather than a real-world deployment.}
        \label{fig:agg}
    \end{figure*}

\section{Summary}

    The FeTS Challenge 2024 envisions FL as a transformative paradigm for multi-institutional collaboration, enabling robust, fair, and generalizable models without compromising patient privacy. The challenge aims to refine FL techniques to improve personalized diagnostics and treatment planning, emphasizing on glioma patients.
    
    The task is focused on advancing aggregation methods for federated consensus models (Figure~\ref{fig:agg}), providing tools that enable replicating an FL environment on a single machine. Using a curated multi-parametric MRI dataset from the BraTS glioma benchmark and clinically relevant segmentation metrics~\citep{maier2024metrics,reinke2024understanding}, the challenge ensures real-world relevance while maintaining compliance with privacy regulations, like HIPAA and GDPR. It addresses critical challenges such as data scarcity and heterogeneity in healthcare AI while offering clear innovation opportunities in areas such as weight aggregation algorithms, hyperparameter tuning, and collaborator selection.

    It is designed for researchers with programming expertise, through low-code tools GaNDLF~\citep{pati2021gandlf} and openFL~\citep{foley2022openfl}, but with clear ``innovation hotspots'' in the code, i.e., lines of the script where innovation is encouraged, including the aggregation algorithm itself, the hyper-parameter tuning, and the selection of collaborators per round. This way innovators/participants can seamlessly integrate new ideas in a code that is reproducible and has already been used successfully for three iterations of a challenge (2021, 2022, and now 2024) that attracted participations from research groups around the globe.

\section{Resource Availability}

    \subsection{Code Location}
    % This section should only include the resource’s webpage address from a permanent public repository. (Note that DOIs are preferred.)
        To enable reproducibility by the scientific community, the instructions and code used for the rankings have been made publicly available\footnote{\url{https://github.com/FETS-AI/Challenge}}. The GaNDLF framework, which acts as the backbone for AI training and development, is also available for public use\footnote{\url{https://github.com/mlcommons/GaNDLF}} \citep{pati2021gandlf}, as well as the backend responsible for the federated learning orchestration (OpenFL by Intel)\footnote{\url{https://github.com/securefederatedai/openfl}}\citep{reina2021openfl,foley2022openfl}.
        
    \subsection{Relevant Research}
    % This section should indicate the potential ML/imaging use cases that the authors envision their resource facilitating and the specific potential clinical domain(s).

        FeTS Challenge 2024 facilitates research on FL through benchmarking algorithms for optimal weight aggregation in FL setups, toward developing models that generalize across diverse clinical datasets without direct data sharing. It also trains research relevant robust AI models to segment glioma subregions in mpMRI data, but bears implications in broader clinical applications (e.g., radiology, cardiology, pathology). Most crucial is the potential of such refined FL techniques to facilitate multi-site collaborations in creating AI models for rare diseases or underserved populations where data availability is limited.

    \subsection{Licensing}
        % Clearly describe the licensing terms. 
        % If there is an existing licence that is followed this could just be a single sentence. Examples of acceptable licenses include but are not limited to CC-BY, CC-BY-NC, Apache v2, MIT.
        
        The FeTS Challenge 2024 and its associated resources (FeTS, GaNDLF, and OpenFL) adhere to an Apache License\footnote{\url{http://www.apache.org/licenses/}}. 
        This ensures that the resources are freely available for research, development, and deployment in various academic and clinical applications.
        
        % \begin{itemize}
        %     \item \textbf{FeTS UI License:} License Type: BSD-style open-source license, compatible with the Open Source Definition by the Open Source Initiative, containing no restrictions on software use\footnote{\url{https://www.med.upenn.edu/cbica/software-agreement.html}}
            
        %     \item \textbf{GANDLF License:} License Type: BSD-style open-source license, compatible with the Open Source Definition by the Open Source Initiative, containing no restrictions on software use\footnote{\url{https://www.med.upenn.edu/cbica/software-agreement.html}}
            
        %     \item \textbf{OpenFL License:} License Type: Apache License, Version 2.0 (January 2004)\footnote{\url{http://www.apache.org/licenses/}}
        % \end{itemize}

\section{Materials \& Methods}

    \subsection{Data}
    
        This challenge leverages data from BraTS 2021~\citep{baid2021rsnaasnrmiccai,bakas2017advancing,bakas2018identifying}, a multi-institutional dataset that has been evolving over the span of a decade, and has supported multiple challenges \citep{menze2014multimodal,bakas2017segmentation_1,bakas2017segmentation_2, adewole2025brats, labella2024analysis, mehta2022qu, labella2024multi, amiruddin2025training, kofler2025brats, adewole2023brain,labella2405brain,maleki2025analysis,labella2024brain, de20242024,kazerooni2024brain,moawad2024brain,kofler2023brain,li2024brain}. BraTS 2021 contains mpMRI scans of glioma patients, routinely acquired during standard clinical practice along with their reference standard annotations for the evaluated tumor subregions. These are augmented with metadata that identify the partitioning of the scans in a de-identified manner. Each patient case contains four structural mpMRI scans at the pre-operative baseline timepoint: i) native T1-weighted (T1) and ii) contrast-enhanced T1 (T1-Gd), iii) T2-weighted (T2), and iv) T2 Fluid Attenuated Inversion Recovery (T2-FLAIR). 
     
         \subsubsection{Data Pre-processing}
     
             The exact pre-processing pipeline applied to all the data considered in the present FeTS challenge is identical with the one evaluated and followed by the BraTS challenge and previous FeTS challenge iterations. Input scans (i.e., T1, T1-Gd, T2, T2-FLAIR) are registered to the same anatomical atlas (i.e., SRI-24 \citep{sri}) using the Greedy diffeomorphic registration algorithm \citep{yushkevich2016fast}, ensuring a common spatial resolution of $(1 mm^3)$. After completion of the registration process, brain extraction is done to remove any apparent non-brain tissue, using a deep learning approach specifically designed for brain MRI scans with apparent diffuse glioma. This algorithm utilizes a novel training mechanism that introduces the brain's shape prior as knowledge to the segmentation algorithm \citep{thakur2020brain}. All pre-processing routines have been made publicly available through the Cancer Imaging Phenomics Toolkit (CaPTk\footnote{\url{https://www.cbica.upenn.edu/captk}}) \citep{captk_1,captk_2,captk_3} and the FeTS tool \citep{pati2022federated}.
         
        \subsubsection{Annotation Protocol}
    
             The skull-stripped scans are annotated to indicate the brain tumor subregions. The annotation process follows a predefined clinically approved annotation protocol that describes the detailed radiologic appearance of each tumor subregion of the MRI scans. In summary, the tumor subregions are: 

            \begin{enumerate}
                \item the enhancing tumor (ET), which delineates the hyperintense signal of the T1-Gd, after excluding the vessels.
                \item the necrotic tumor core (NCR), which outlines regions appearing dark in both T1 and T1-Gd images (denoting necrosis/cysts), and dark regions in T1-Gd and bright in T1.
                \item the tumor core (TC), which includes the ET and NCR, and represents what is typically resected during a surgical operation.
                \item the whole tumor (WT), which includes the peritumoral edematous and infiltrated tissue (ED), delineates the regions characterized by the hyperintense abnormal signal envelope on the T2-FLAIR sequence.
            \end{enumerate}
            % The provided segmentation labels have values of $1$ for NCR, $2$ for ED, $4$ for ET, and $0$ for everything else.
            
            During its collection, BraTS followed a strict peer-review process for quality control, where each case was assigned to pairs of annotator-approvers. Annotator experience ranged across various levels of clinical / academic ranks, while approvers were the two experienced board-certified neuroradiologists (with $\geq$ 13 years of glioma experience). The annotators were given flexibility on which tool to use, and whether to follow a complete manual annotation approach, or a hybrid one with automated initial annotations followed by their manual refinements. Afterward, the produced annotations were passed to the corresponding approver, who evaluated them in tandem with the original mpMRI, and either signs them off or, in case of quality issues, returned them to the annotators for refinements. This iterative approach is followed for all cases until their respective annotations reaches satisfactory quality for publication, and noted as final reference standard segmentation labels.

            \begin{table}[t]
                \setlength{\tabcolsep}{5pt}
                \centering
              \caption{Overview of case numbers in training, validation, and test sets. Datasets with source \textit{BraTS'21} are centralized. $^*$based on partitioning 1/2}
                \begin{tabular}{|l | c  c  c|} 
                 \hline
                  & Training & Validation & Test \\ [0.5ex] 
                 \hline
                 \# cases & 1251 & 219 & 570 \\ 
                 \# institutions & 23/33$^*$ & 10 & 12 \\
                 Source & BraTS'21 & BraTS'21 & BraTS'21 \\
                 \hline
                \end{tabular}
                \label{tab:case_numbers}
            \end{table}
            \begin{figure*}[tb]
            \centering
            \includegraphics[width=\textwidth]{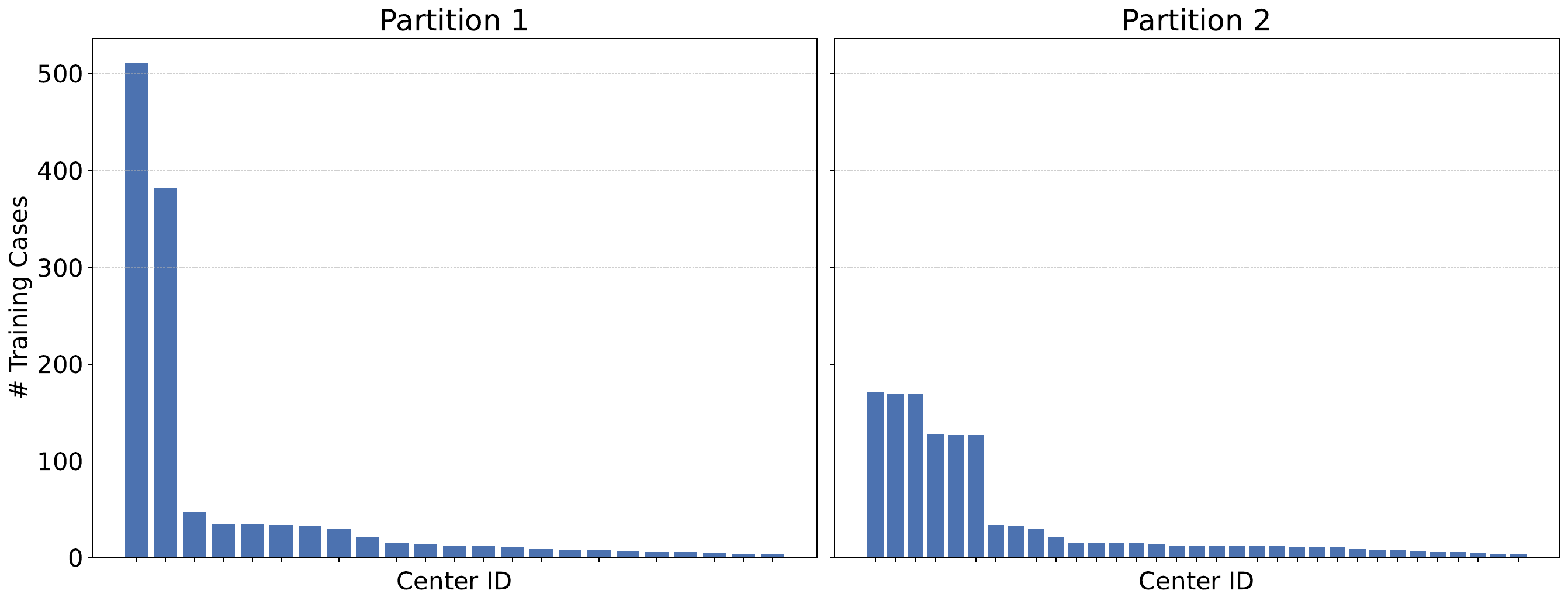}
            \caption{Official partitionings of the training set.}
            \label{fig:train_partitionings}
        \end{figure*}
        \subsubsection{Training, Validation, and Test case characteristics}
    
            Of the original BraTS 2021 dataset, only the subset of radiographically visible glioma is included in the FeTS challenge, while cases without apparent enhancement are excluded. The exact numbers can be found in Table \ref{tab:case_numbers}. Training cases encompass the mpMRI volumes, the corresponding tumor subregion annotations, as well as a pseudoidentifier of the site where the scans are acquired. Validation cases, however, only contain the mpMRI volumes, without any accompanying reference standard annotations or site pseudo-identifiers. We explicitly provide two schemas to partition the provided data (Figure \ref{fig:train_partitionings}):
             % \begin{enumerate}
             %     \item Natural geographical partitioning by institution (partitioning 1)
             %     \item Artificial partitioning using imaging information (partitioning 2), by further sub-dividing each of the 5 largest institutions in partition 1 into two parts according to the whole tumor size (median size was used as threshold).
             % \end{enumerate}

             \begin{enumerate}
                \item Natural geographical partitioning by institution (partitioning 1, 23 sites)
                \item Artificial partitioning using imaging information (partitioning 2, 33 sites), by further sub-dividing each of the 5 largest institutions in partition 1 into three parts after sorting samples by their whole tumor size.
            \end{enumerate}
            80\% of the dataset is kept for federated training (train and validation split), while the rest is set aside as a test set to evaluate the performance of submitted algorithms. The test set is never shared with the challenge participants.

\subsection{Software Stack}
\label{sec:task1}

\subsubsection{Model Architecture}
    \label{sec:arch}
    As the challenge focuses on the development of aggregation methods, the architecture of the segmentation model itself remains fixed across all participants. Following current literature and previous iterations of the challenge, the segmentation model chosen for this role is U-Net \citep{unet} as the architecture has consistently performed well on medical imaging datasets \citep{isensee2018nnu,thakur2020brain,drozdzal2016importance,he2016deep,cciccek20163d,pati2021gandlf}. The U-Net architecture is composed of an encoder-decoder architecture, where the encoder consists of layers performing convolutions and downsampling, while the decoder consists of layers performing transpose-convolution and upsampling. It includes skip connections in every convolution block---i.e., concatenated feature maps paired across the encoder and the decoder layer to improve context and feature re-usability, capturing information at multiple scales/resolutions.

    \subsubsection{Federated Training}
    \label{sec:aggregation}
    An infrastructure for federated tumor segmentation has been provided to all participants indicating the spots on the code where they are expected to make changes (``innovation hotspots''). The primary objective is to develop methods for effective aggregation of local segmentation model weight updates, given the partitioning of the data into their real-world distribution.
    
    The aggregation mechanism follows extensive prior literature \citep{sheller2018multi,sheller2020federated,isensee2018nnu,pati2021gandlf}, and is illustrated by Figure \ref{fig:agg}. Models are trained locally in each individual site and sent back to the aggregator at the end of each FL round. At the start of each round, each collaborator locally validates any model it receives from the central aggregation server (\textit{aggregator}), then trains the model received from the \textit{aggregator} on their local data. The local validation results along with the model updates are then returned to the \textit{aggregator}, which combines model updates from all sites to produce a new consensus model. The consensus model is then passed back to each \textit{collaborator}, and starts a new federated round.

    % The ranking process evaluates team performance of teams based on key metrics derived from their predictions, including Dice coefficients and Hausdorff distances for all three tumor subregions (ET, TC, WT). Teams are ranked separately for each metric on a per-sample scale: Dice metrics are ranked in descending order (higher is better), while Hausdorff metrics are ranked in ascending order (lower is better). These rankings are calculated using a group-wise ranking approach, and tied scores are averaged using the "average" tie resolution method. The rankings across all metrics are then aggregated into a cumulative rank by averaging the ranks for each team across all cases and metrics, resulting in a score that represents the team’s overall performance.
    
    % To further refine the ranking, a communication efficiency metric is incorporated, which measures aspects like the projected Dice over time. This communication score is ranked (with higher values ranked better) and weighted based on a predefined factor before being added to the cumulative rank. The final rankings, with and without communication scores, are visualized using violin plots, which display the distribution of cumulative ranks for each team, overlaid with boxplots showing medians and interquartile ranges. The X-axis of these plots represents the cumulative rank scores, where lower values indicate better performance, and the Y-axis lists the competing teams. This ranking system ensures a holistic evaluation of performance by considering prediction quality alongside communication efficiency.

\section{Quantitative Performance Evaluation}

        \label{sec:eval_metrics}
        
        Participants are called to produce segmentation labels of the different glioma subregions (ET, TC, WT). For each region, the predicted segmentation is compared with the reference standard segmentation using the following metrics:

         \subsection{Dice similarity coefficient (DSC)}
            The DSC is a metric commonly used to evaluate the performance of segmentation tasks. It measures the extent of spatial overlap of predicted masks ($PM$) and the provided reference standard ($RS$), while considering their union, thereby addressing both over-segmentation and under-segmentation. It is defined as
         
            \begin{equation}
            \label{eq:dice}
                DSC = \frac{2|RS \cap PM|}{|RS|+|PM|} \, .
            \end{equation}

         \subsection{Hausdorff distance (HD)}
            This metric quantifies the distance between the boundaries of the reference standard labels against the predicted labels. It originates from set theory and measures the maximum distance of a point set to the nearest point in another set \citep{hausdorff}. This makes the HD sensitive to local differences, as opposed to the DSC, which represents a global measure of overlap. For the specific problem of brain tumor segmentation, local differences are important for properly assessing the quality of the segmentation. In this challenge, the 95\textsuperscript{th} percentile of the HD between the contours of the two segmentation masks is calculated, which is a variant of HD that is more robust to outlier pixels:
                \begin{align}
                \label{eq:haus}
                HD_{95}(PM, GT) &= \max \Big\{ 
                \underset{p \in PM}{P_{95\%}}\, d(p, GT), \nonumber \\
                &\quad\quad\quad \underset{g \in GT}{P_{95\%}}\, d(g, PM) \Big\},
                \end{align}
            where $d(x, Y) = \min_{y\in Y} ||x - y||$ is the distance of $x$ to set $Y$.

            % \subsubsection*{Communication cost during model training}
            % The ``cost'' represents the budget time, which is the product of number of bytes sent/received multiplied by number of federated rounds. This metric was used to assess the performance of the weight aggregation methods. In the 2021 instance, this cost was weighted the same as $DSC$ and $HD_{95}$ from each of the regions of interest.
        \subsection{Convergence Score}
            % Communication efficiency is the metric conceived to estimate the efficiency of the model and encompasses the time taken to train and evaluate as well as the communication costs (download and upload of model weights between each collaborator and the central server). To calculate this metric, we simulate the time taken to run each round, and normalize across a single hardware system to assure the final scoring is not impacted by hardware quality. Model download and upload times are computed based on the real-world timings measured during the FeTS initiative \citep{pati_federated_2022}. The communication efficiency metric is computed as the area under the validation learning curve over 1 week of simulated time (higher is better). The horizontal axis measures simulated runtime and the vertical axis measures the current best score, computed as the average of enhancing tumor, tumor core, and whole tumor DSC scores over the validation split of the training data.

            Convergence Score is used to estimate the efficiency of the model and encompasses the time taken to train and evaluate as well as the communication costs (download and upload of model weights between each collaborator and the central server).
            
            We simulate the cumulative time taken per round, breaking it down to four components: training time $T_{\mathrm{train}}$, validation time $T_{\mathrm{val}}$, model weight download $T_{\mathrm{down}}$ and upload time $T_{\mathrm{up}}$. In each round, the simulated time per collaborator $k$ is
                \begin{equation}
                    T_k = T_{\mathrm{down},k} + T_{\mathrm{up},k} + T_{\mathrm{val},k} \cdot N_{\mathrm{val},k} + T_{\mathrm{train},k} \cdot N_{\mathrm{train},k}
            \end{equation}
            The total time for each round is $\max_k \{T_k\}$. To simulate a realistic FL setup, $T_{\mathrm{x},k}$ was sampled from a normal distribution: $T_{\mathrm{x},k} \sim \mathcal{N}(\mu_{\mathrm{x},k}, \sigma_{\mathrm{x},k})$, where x can be replaced with train/val/down/up. The parameters of the normal distribution are fixed but different for each client $k$, and based on time measurements derived from the largest to-date real-world FL study, which used the same model---the FeTS initiative \citep{pati_federated_2022}.
            % Random seeds guarantee that these are identical for all FL experiments, so that all participants use the same timings.

            In each federated round we compute the mean DSC on a fixed split (20\%) of the  training data and the simulated round time $T$. Over the course of an experiment, this results in a DSC-over-time curve.
            % The validation DSC can in some cases decrease at later times (e.g. due to overfitting or randomness in the optimization), but as the model with the best DSC is used as the final model, such a decrease should not be penalized. Therefore,
            A projected DSC curve is computed as $\mathrm{DSC_{proj}} (t) = \max_{t^\prime \leq t}\mathrm{DSC(t^\prime)}$. The convergence score metric is calculated as the area under that projected DSC-over-time curve:

            \begin{equation}
            S_{\text{conv}} = \int_{t_0}^{T} \mathrm{DSC_{proj}}(t) \, dt
            \end{equation}
            
            where:
            \begin{itemize}
                \item $S_{\text{conv}}$ is the convergence score. Higher values of this metric indicate enhanced convergence and thus a superior FL approach in terms of efficiency.
                \item $\mathrm{DSC_{proj}}(t)$ is the projected DSC at time $t$.
                \item The integral runs from $t_0$ (starting time) to $T$ (final time).
            \end{itemize}
            
            To standardize the time-axis for the convergence score among participants, all FL experiments performed during the challenge are limited to one week of simulated total time, which was a realistically feasible duration based on the experience from the FeTS initiative \citep{pati_federated_2022}. The FL runs were terminated once the simulated time exceeds one week and the model with the highest validation score before the last round is stored, to assure that a long last round exceeding the time limit does not benefit the participant.

        \subsection{Ranking Strategy}
            Before evaluating the submissions on the test set, algorithms are re-trained by the organizers, to ensure reproducible results and to prevent data leakage between federated sites. To standardize comparisons, the rankings are computed independently for each test case before aggregating them across the dataset, so performance is assessed fairly across different cases without being dominated by any single case. DSC scores, which measure the overlap between predicted and reference standard segmentations, are ranked in descending order since higher values indicate better segmentation performance. Conversely, Hausdorff distances, which quantify boundary errors, are ranked in ascending order as lower values correspond to more precise boundary delineations. 
                    
            A tie-breaking strategy using rank averaging is applied to ensure consistent rankings when multiple teams achieve the same performance on a given test case. Specifically, if $k$ teams have the same score for a particular metric, they all receive the average rank they would occupy if ranked distinctly. For example, if three teams tie for the second-best score, instead of being arbitrarily assigned ranks 2, 3, and 4, they are each assigned the average rank $(2+3+4)/3 = 3$. This method prevents unfair advantages or disadvantages due to arbitrary rank assignment and ensures a smooth aggregation of ranks across multiple cases. The overall ranking score for each team is then computed as the mean of its assigned ranks across all Dice and Hausdorff metrics, leading to a comprehensive performance measure.
            
            Beyond segmentation quality, the framework integrates a communication efficiency score to refine the rankings. The communication metric, which reflects the efficiency of model convergence during training, is ranked separately using the same averaging-based ranking strategy. Since a higher communication score indicates better efficiency, rankings for this metric are assigned in descending order. This additional ranking is incorporated into the final ranking score using a predefined weight $w$, ensuring that teams that achieve competitive segmentation performance with lower communication costs are favored. The cumulative ranking score, incorporating segmentation and communication efficiency, is computed as the mean of all individual rankings (Dice, Hausdorff, and convergence score). 
            
            \begin{table*}[t!]
            \setlength{\tabcolsep}{7pt}
            \centering
            \caption{Teams and the corresponding papers of their submissions. Note that HTTUAS submitted 2 papers, proposing two distinct methods abbreviated in the text based on the technique they're deploying (Rec for Recommender Engine and RL for Reinforcement Learning).}
            \begin{tabular}{l p{16cm}}
            \hline
            Team & Method (Paper Title) \\
            \hline
            \textcolor{SNU}{SNU} & FedPOD: the deployable units of training for federated learning \\ 
            \hline
            \textcolor{rigg}{rigg} & FedPID, an aggregation method for Federated Learning \citep{machler2024fedpid}\\ 
            \hline
            \textcolor{HTTUAS-REC}{HTTUAS (Rec)} & Recommender Engine for Client Selection in Federated Brain Tumor Segmentation \citep{khan2024recommender}\\ 
            \hline
            \textcolor{HTTUAS-RL}{HTTUAS (RL)} & Election of Collaborators via Reinforcement Learning for Federated Brain Tumor Segmentation \citep{khan2024election}\\ 
            \hline
            \textcolor{Flair}{Flair} & Adaptive Federated Learning for Brain Tumor Segmentation: A Clustered Approach with Bias-Variance Balancing \\ 
            \hline
            \textcolor{ReMIC}{ReMIC} & Federated Tick-Tack \\ 
            \hline
            \end{tabular}
            \label{tab:teams_description}
            \end{table*}

            Mathematically, the final ranking score $R_t$ for team $t$ is given by:
            
            \begin{equation}
            R_t = \frac{1}{N + w} \sum_{m \in M} r_{t,m} + \frac{w}{N + w} r_{t,comm}
            \label{eq:ranking_formula}
            \end{equation}

            where:
            
            \begin{itemize}
                \item $M$ is the set of segmentation metrics (Dice and Hausdorff) evaluated for each tumor subregion (ET, TC, WT).
                \item $N = |M|$ is the total number of segmentation rankings per test case, calculated as the number of metrics (2: Dice and Hausdorff) multiplied by the number of tumor subregions (3: ET, TC, WT), giving $N = 2 \times 3 = 6$,
                \item $r_{t,m}$ is the rank of team $t$ for metric $m$,
                \item $r_{t,comm}$ is the rank of team $t$ based on communication efficiency,
                \item $w$ is the weight assigned to the communication metric.
            \end{itemize}

            This formulation ensures a balanced evaluation of segmentation accuracy and computational efficiency, where the influence of communication efficiency is modulated by the chosen weight $w$. When $w = 0$, rankings depend solely on segmentation performance, whereas increasing $w$ gives more importance to communication efficiency.

    \section{Limitations}
        As a simulation-based setup, the FeTS 2024 challenge comes with limitations. While we approximate weight download and upload times in our convergence score metric, it does not fully reflect the communication heterogeneity across sites, which includes variable bandwidth, latency, dropped connections, and asynchronous updates. Second, differences in hardware availability (e.g., GPU memory, CPU load, and system failures) are abstracted away, but these factors strongly influence performance in real-world studies. Finally, security protocols, and the operational complexities of coordinating institutions are not represented here.
        
        Despite these limitations, a simulated setup ensures reproducibility, fairness, and scalability for benchmarking innovations in aggregation methods. Such simulations are necessary prior to real-world deployments as they assess multiple promising aggregation methods without the expensive nature of actual large-scale coordination. They are essentially a preliminary step to future work that would bridge simulation and deployment, incorporating realistic communication and hardware variability into benchmarking protocols.

    \section{Participating Methods}

        A total of 6 submissions by 5 teams from four continents were submitted to FeTS Challenge 2024. The six proposed algorithms are outlined in table~\ref{tab:teams_description} and their innovations are as follows: 
        
        \textbf{\tikz\draw[ReMIC, fill=ReMIC] (0,0) circle (.5ex); \, Federated Tick-Tack by ReMIC}: Fed Tick-Tack presents a novel approach to federated learning by introducing a two-phase aggregation technique designed to enhance model robustness and accuracy, especially in heterogeneous data environments. Instead of updating models at the end of each communication round, Fed Tick-Tack alternates between two distinct phases: Tick and Tack. In the Tick phase, an aggregated model is distributed to selected collaborators, who then locally train and return their models, weighted according to their individual importance (i.e. a weight assigned to each collaborator that reflects their model's influence on the overall aggregated model). The Tack phase focuses on updating these weights by calculating the differences between consecutive model proposals. This iterative adjustment ensures that the final model not only reflects the most recent learning but also adapts to the individual progress of each collaborator. Furthermore, Fed Tick-Tack supports both scalar and parameter-specific weight adjustments, offering flexibility in how collaborators' contributions are integrated. This method also introduces ranked batches to organize collaborators based on performance, toward balanced and efficient training rounds.
        
        \textbf{\tikz\draw[Flair, fill=Flair] (0,0) circle (.5ex); \, Clustered Approach with Bias-Variance Balancing by Flair}: Flair's approach integrates client selection, hyper-parameter tuning, and aggregation methods to enhance federated model training. The method involves clustering clients using k-means based on their training times, which minimizes idle times and ensures that clients with smaller datasets are effectively utilized. An adaptive strategy is used for hyper-parameter selection, including dynamic adjustment of local epochs and a modified cosine annealing schedule for learning rates. The aggregation method improves upon traditional FedAvg by incorporating Validation Loss Ratio (VLR) and an Overfit Penalty (OFP) to balance contributions based on validation performance and to address overfitting.
        
        \textbf{\tikz\draw[HTTUAS-REC, fill=HTTUAS-REC] (0,0) circle (.5ex); \, Recommender Engine for Client Selection by HTTUAS}: The study introduces a novel client selection protocol for Federated Learning in brain tumor segmentation, leveraging a recommender engine based on Non-Negative Matrix Factorization (NNMF) combined with a hybrid content-based and collaborative filtering approach. The NNMF decomposes historical performance metrics to identify suitable collaborators, while a fallback mechanism ensures continued operation in the absence of sufficient data. Additionally, this method presents Harmonic Similarity Weighted Aggregation (HSimAgg), an enhancement of the SimAgg \citep{khan2021adaptive} algorithm that uses harmonic mean aggregation to robustly handle outliers and extreme values, improving the accuracy and reliability of the federated model.
        
        \textbf{\tikz\draw[HTTUAS-RL, fill=HTTUAS-RL] (0,0) circle (.5ex); \, Election of Collaborators via Reinforcement Learning by HTTUAS} and similarity-weighted aggregation (SimAgg) \citep{khan2021adaptive} approach designed to optimize collaborator selection in federated brain tumor segmentation. The method employs multi-armed bandit algorithms, specifically Epsilon-greedy (EG) and Upper Confidence Bound (UCB), to manage the selection of collaborators and enhance model generalization. RL-HSimAgg balances exploration and exploitation to promote effective training across diverse datasets by dynamically choosing collaborators based on their performance. The approach also incorporates a similarity-weighted aggregation method to handle outliers by using harmonic mean, thereby improving robustness in FL environments.
        
        \textbf{\tikz\draw[rigg, fill=rigg] (0,0) circle (.5ex); \, FedPID by rigg}: Building on their previous participations FedCostWAvg and FedPIDAvg \citep{machler2021fedcostwavg, machler2022fedpidavg}, FedPID refines the aggregation strategy by incorporating improvements in how the integral term is computed. Unlike FedPIDAvg, which used a simple integration of the loss function, FedPID measures the global drop in cost since the first round. This method integrates a weighted averaging scheme combining dataset sizes, recent cost reductions, and global cost changes to update the model. Additionally, FedPID addresses varying dataset sizes by modeling them with a Poisson distribution, adjusting training iterations accordingly. This approach aims to enhance model performance by balancing local improvements with global progress and handling dataset size variability effectively.
        
        \textbf{\tikz\draw[SNU, fill=SNU] (0,0) circle (.5ex); \, FedPOD by SNU}: FedPOD also builds upon the foundations of FedPIDAvg \citep{machler2022fedpidavg}, to optimize both learning efficiency and communication costs in federated learning. FedPOD complements FedPIDAvg in two ways: (a) it includes outlier nodes that would otherwise be excluded and (b) eliminates the need for historical participant data. Due to these modules, FedPOD aims to better handle skewed data distributions and participant variability. Also, FedPOD is designed to work with Kubernetes' POD units, allowing for dynamic scaling of computational resources through Kubernetes' auto-scaling functionality.

    \section{Results}

        All teams are ranked according to Equation~\eqref{eq:ranking_formula}. For each of the $570$ testing subjects, three tumor regions---ET, TC, and WT---are evaluated using two segmentation measures: the Dice Similarity Coefficient ($DSC$) and the 95th percentile Hausdorff Distance ($HD_{95}$). This results in a total of $570 \times 3 \times 2 = 3,420$ individual rankings across all cases and metrics. As each of these metrics is accounted for three times (one for each modality), we also incorporate convergence score into the ranking process with a weighting factor of $w = 3$. This results in an adjusted total of $570 \times 3 \times 3 = 5,130$ rankings summed per team.

        To provide an overview of segmentation performance, Table~\ref{tab:mean_dice_hd95} presents the mean DSC and HD95 scores for each team. Higher DSC values indicate better segmentation accuracy, whereas lower HD95 values signify more precise boundary delineation.
        
        FedPID and FedPOD methods presented by rigg and SNU respectively are top performers on DSC across all tumor types, indicating superior segmentation accuracy and consistency. The two methods also achieve competitive median Hausdorff Distance at 95th percentile (HD95) scores, demonstrating robustness in handling extreme boundary cases. The best DSC performance is achieved by SNU, with values of 0.733 (ET), 0.751 (WT), and 0.761 (TC). The best HD95 scores (i.e., lowest values) are more distributed: rigg achieves the lowest HD95 for ET (32.246 mm) and TC (31.705 mm), while HTTUAS (Rec) attains the lowest for WT (28.228 mm). These results demonstrate that no single method consistently outperforms across all metrics, highlighting trade-offs between segmentation accuracy and robustness.

                \begin{table}[h]
        \setlength{\tabcolsep}{3pt}
        \centering
        \caption{Mean DSC and HD95 for each team in the FeTS Challenge 2024. The highest DSC values (higher is better) and the lowest HD95 values (lower is better) are highlighted in bold.}
        \begin{tabular}{c|c c c|c c c}
        \hline
            \multirow{2}{*}{\footnotesize{\textbf{Team}}} & \multicolumn{3}{c|}{\footnotesize{\textbf{DSC (↑)}}} & \multicolumn{3}{c}{\footnotesize{\textbf{HD95 (↓)}}} \\
            & \footnotesize{\textbf{ET}} & \footnotesize{\textbf{WT}} & \footnotesize{\textbf{TC}} & \footnotesize{\textbf{ET}} & \footnotesize{\textbf{WT}} & \footnotesize{\textbf{TC}} \\
            \hline
            \footnotesize{\textcolor{SNU}{{SNU}}} & \footnotesize{\textbf{0.733}} & \footnotesize{0.751} & \footnotesize{\textbf{0.761}} & \footnotesize{33.922} & \footnotesize{32.309} & \footnotesize{33.623} \\
            \footnotesize{\textcolor{rigg}{{rigg}}} & \footnotesize{0.722} & \footnotesize{\textbf{0.754}} & \footnotesize{0.748} & \footnotesize{\textbf{32.246}} & \footnotesize{31.122} & \footnotesize{\textbf{31.705}} \\
            \footnotesize{\textcolor{HTTUAS-REC}{{HTTUAS (Rec)}}} & \footnotesize{0.682} & \footnotesize{0.738} & \footnotesize{0.716} & \footnotesize{34.023} & \footnotesize{\textbf{28.228}} & \footnotesize{32.911} \\
            \footnotesize{\textcolor{HTTUAS-RL}{{HTTUAS (RL)}}} & \footnotesize{0.668} & \footnotesize{0.702} & \footnotesize{0.699} & \footnotesize{32.930} & \footnotesize{28.991} & \footnotesize{31.372} \\
            \footnotesize{\textcolor{Flair}{{Flair}}} & \footnotesize{0.658} & \footnotesize{0.651} & \footnotesize{0.681} & \footnotesize{42.637} & \footnotesize{27.893} & \footnotesize{44.622} \\
            \footnotesize{\textcolor{ReMIC}{{ReMIC}}} & \footnotesize{0.620} & \footnotesize{0.645} & \footnotesize{0.644} & \footnotesize{45.724} & \footnotesize{29.030} & \footnotesize{46.426} \\
        \hline
        \end{tabular}
        \label{tab:mean_dice_hd95}
        \end{table}
        
\begin{figure*}[h!]
        \centering
        \begin{minipage}{0.5\textwidth}
            \centering
            \subfloat[Dice ET]{\includegraphics[width=\textwidth]{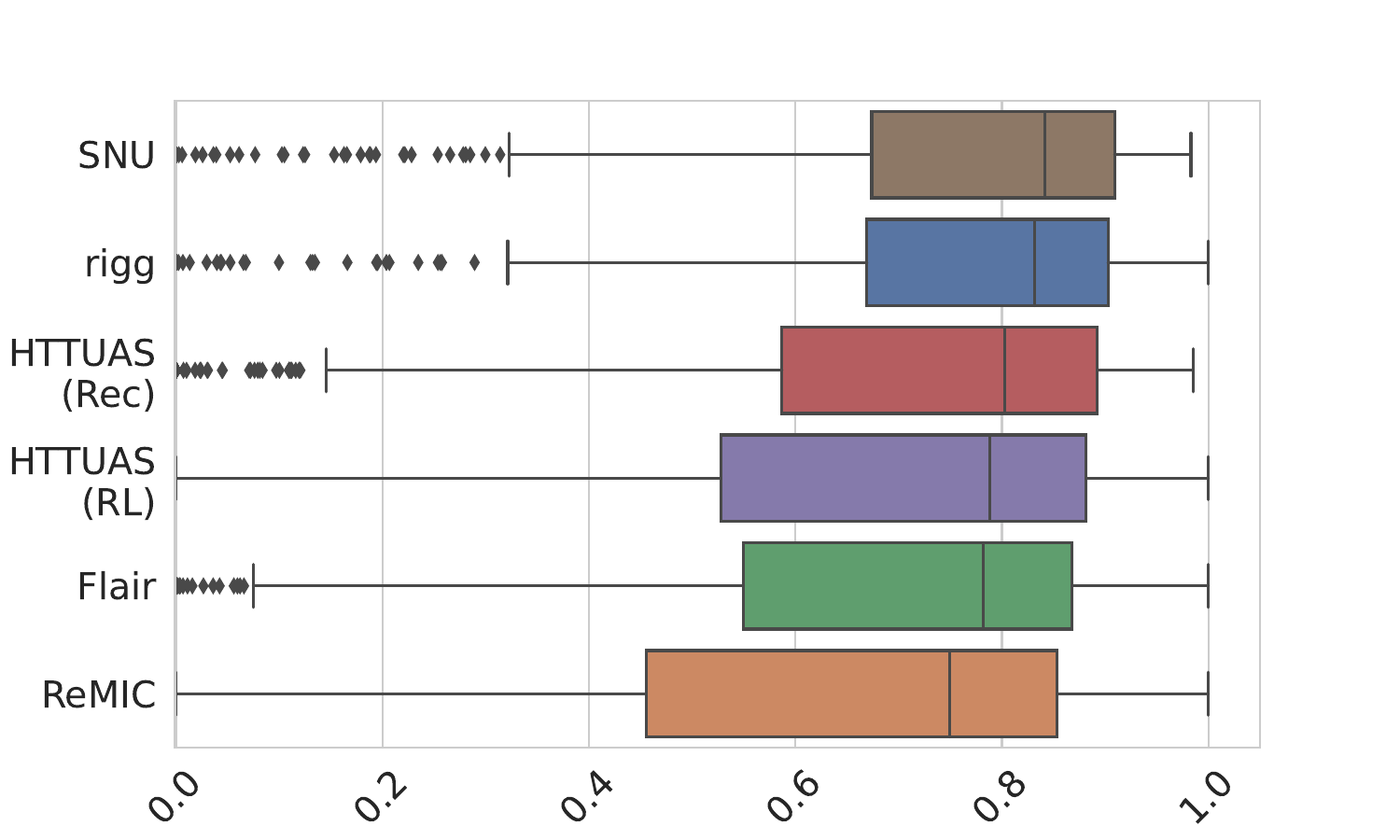}\label{fig:dice_et}}\\[-0.2ex]
            \subfloat[Dice WT]{\includegraphics[width=\textwidth]{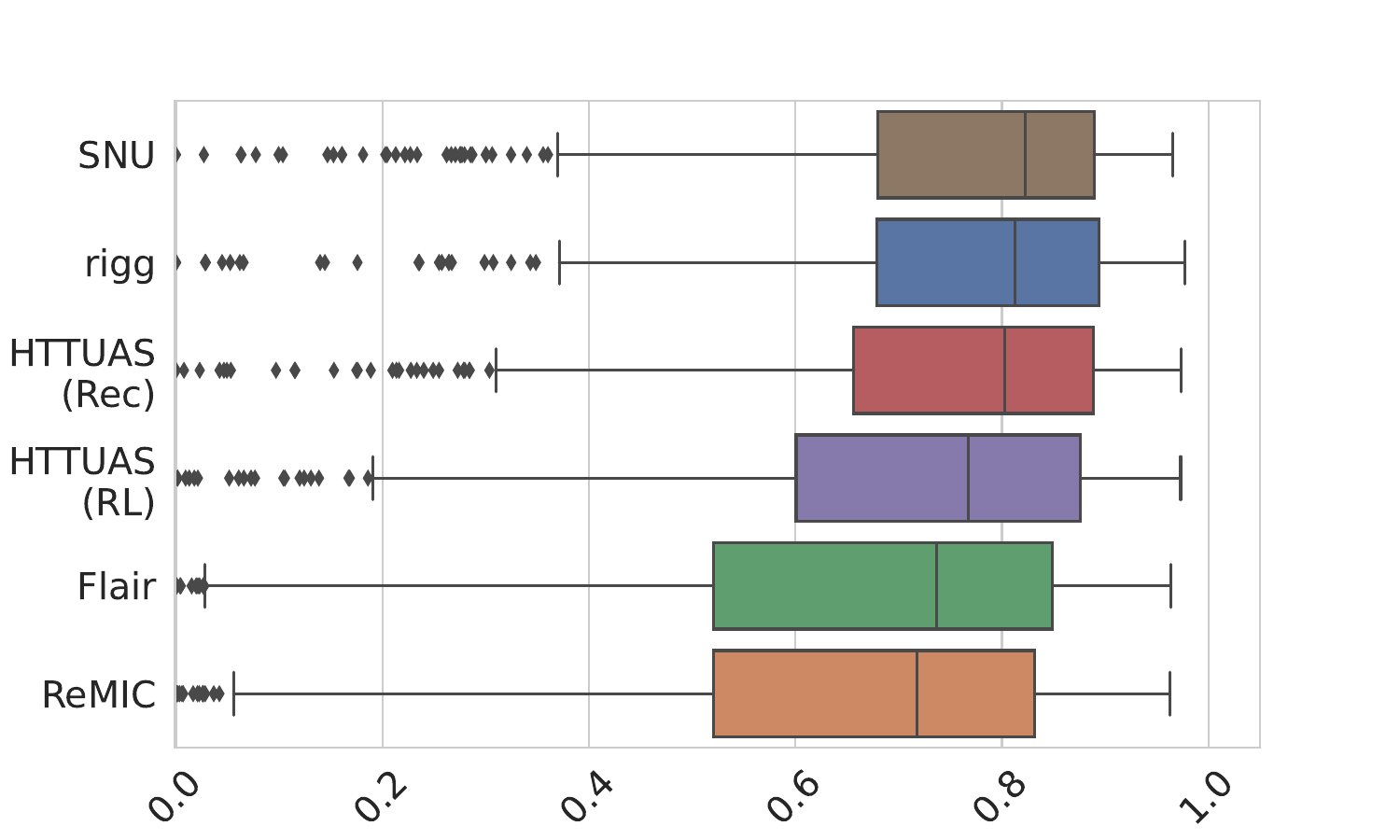}\label{fig:dice_wt}}\\[-0.2ex]
            \subfloat[Dice TC]{\includegraphics[width=\textwidth]{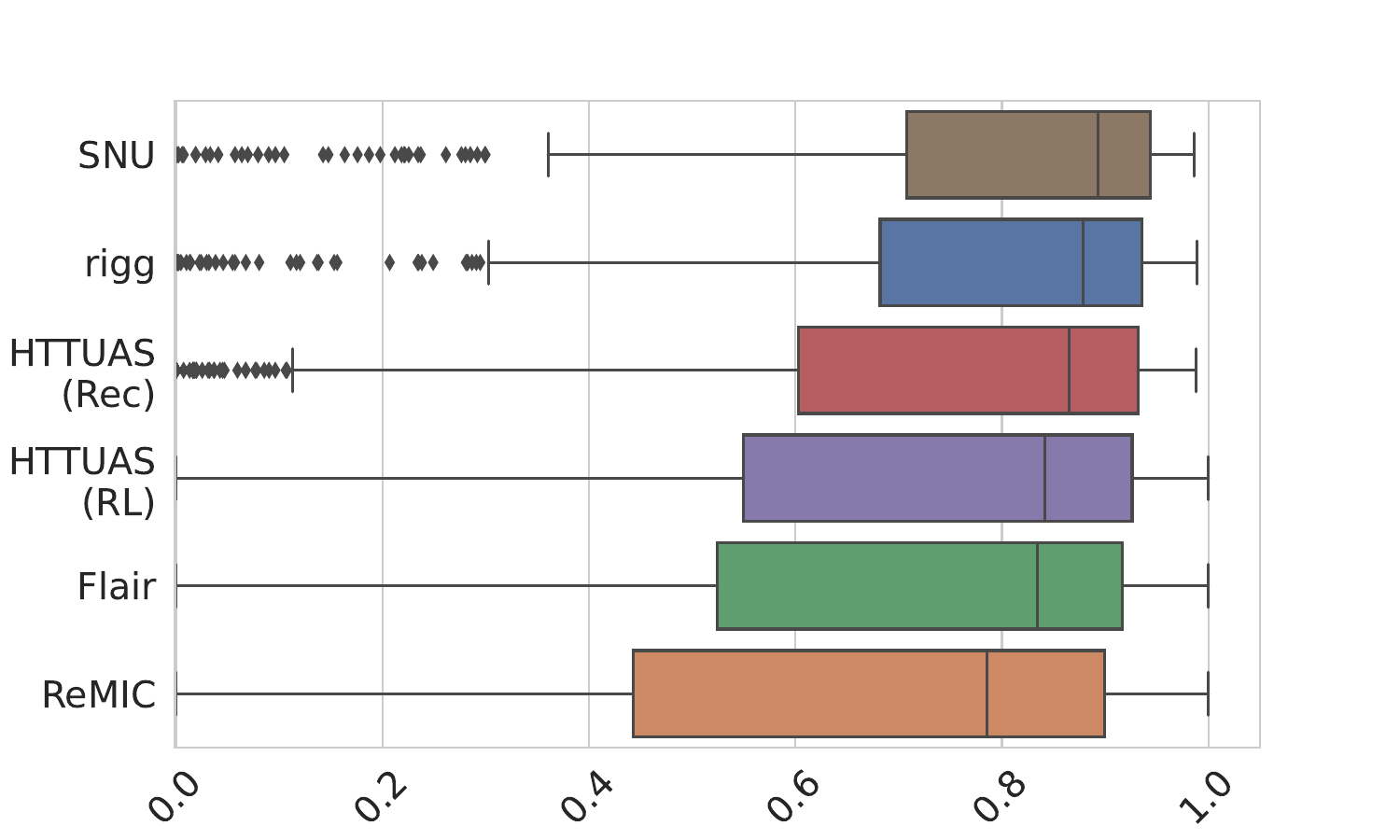}\label{fig:dice_tc}}
        \end{minipage}\hfill
        \begin{minipage}{0.5\textwidth}
            \centering
            \subfloat[HD95 ET]{\includegraphics[width=\textwidth]{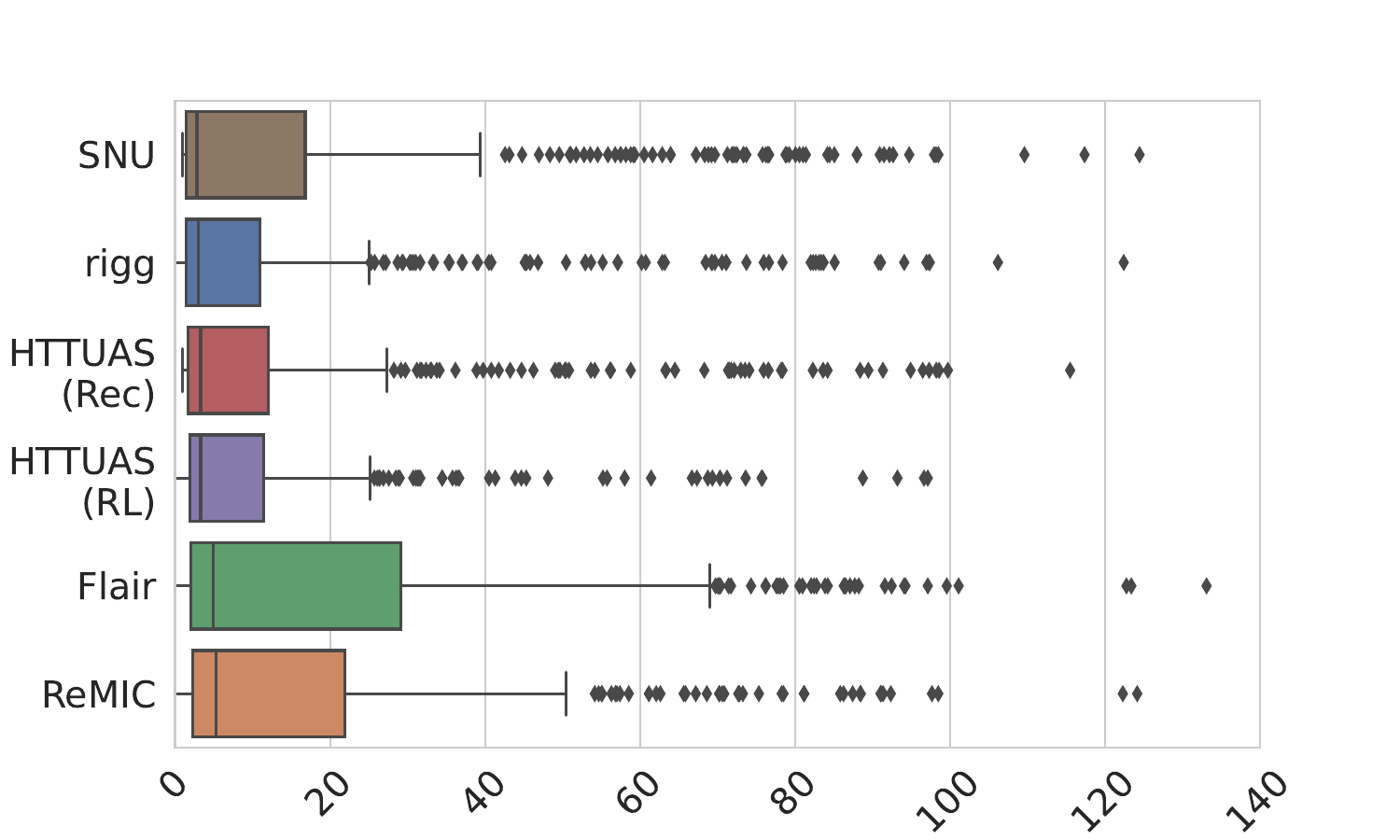}\label{fig:hd95_et}}\\[-0.2ex]
            \subfloat[HD95 WT]{\includegraphics[width=\textwidth]{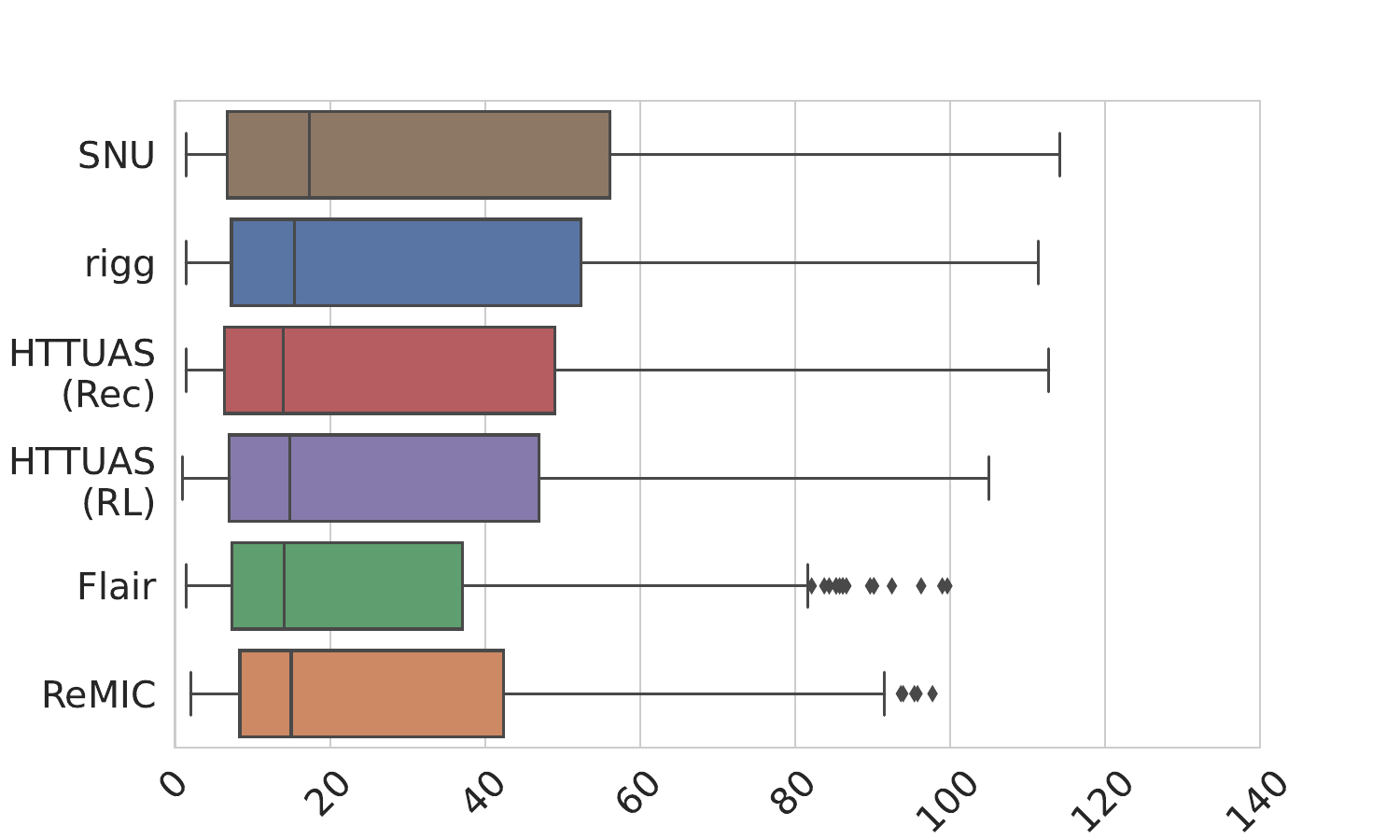}\label{fig:hd95_wt}}\\[-0.2ex]
            \subfloat[HD95 TC]{\includegraphics[width=\textwidth]{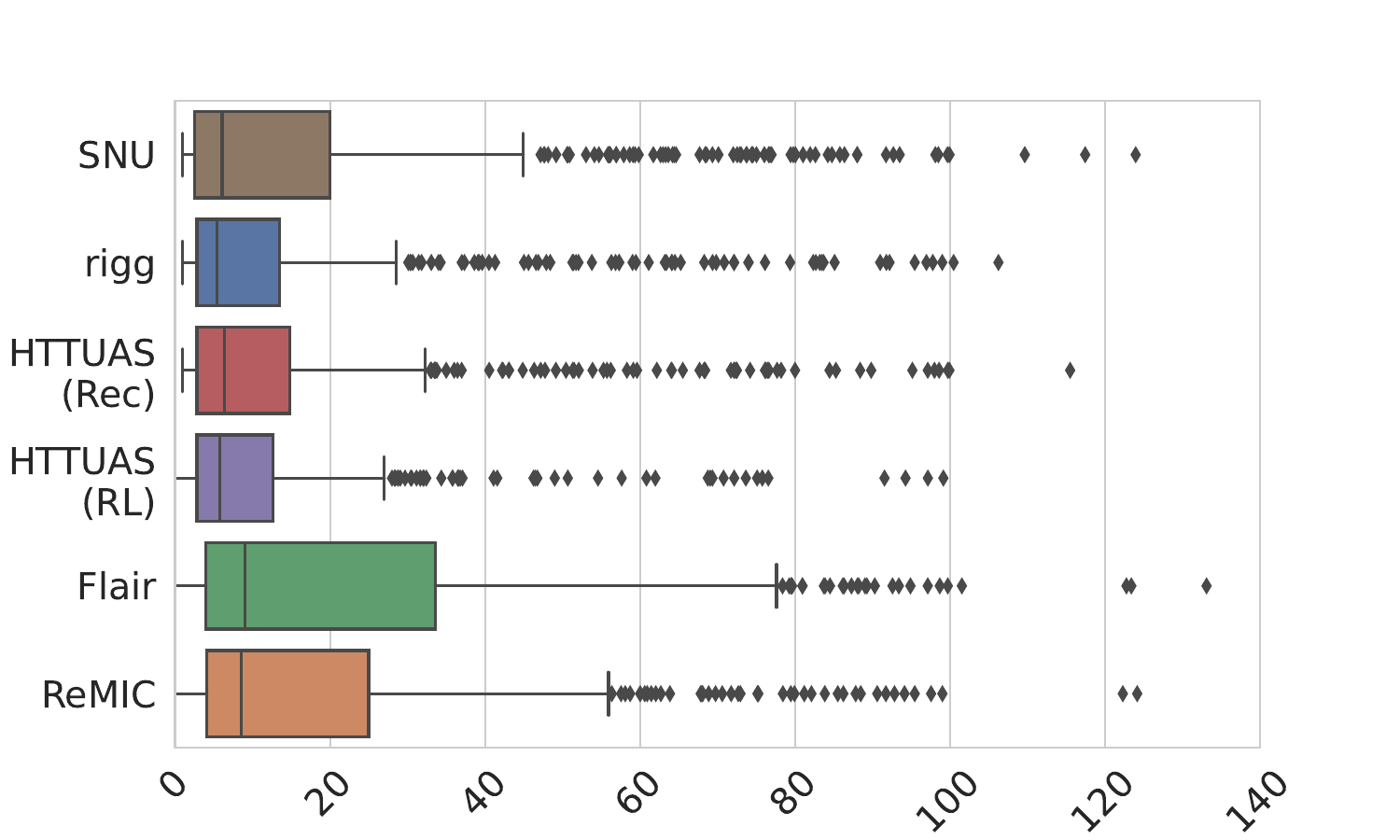}\label{fig:hd95_tc}}
        \end{minipage}
        
        \caption{Box plots for Dice and HD95 metrics, illustrating the span of segmentation performance across different participating methods.}
        \label{fig:combined_dsc_hd}
        \end{figure*}

        Visualizing the distribution of the performance, we also observe substantial variability in performance across different samples, with rigg and SNU exhibiting the lowest variance (Figure ~\ref{fig:combined_dsc_hd}).
                
\begin{figure}[h!]
        \centering
        \includegraphics[width=0.49\textwidth]{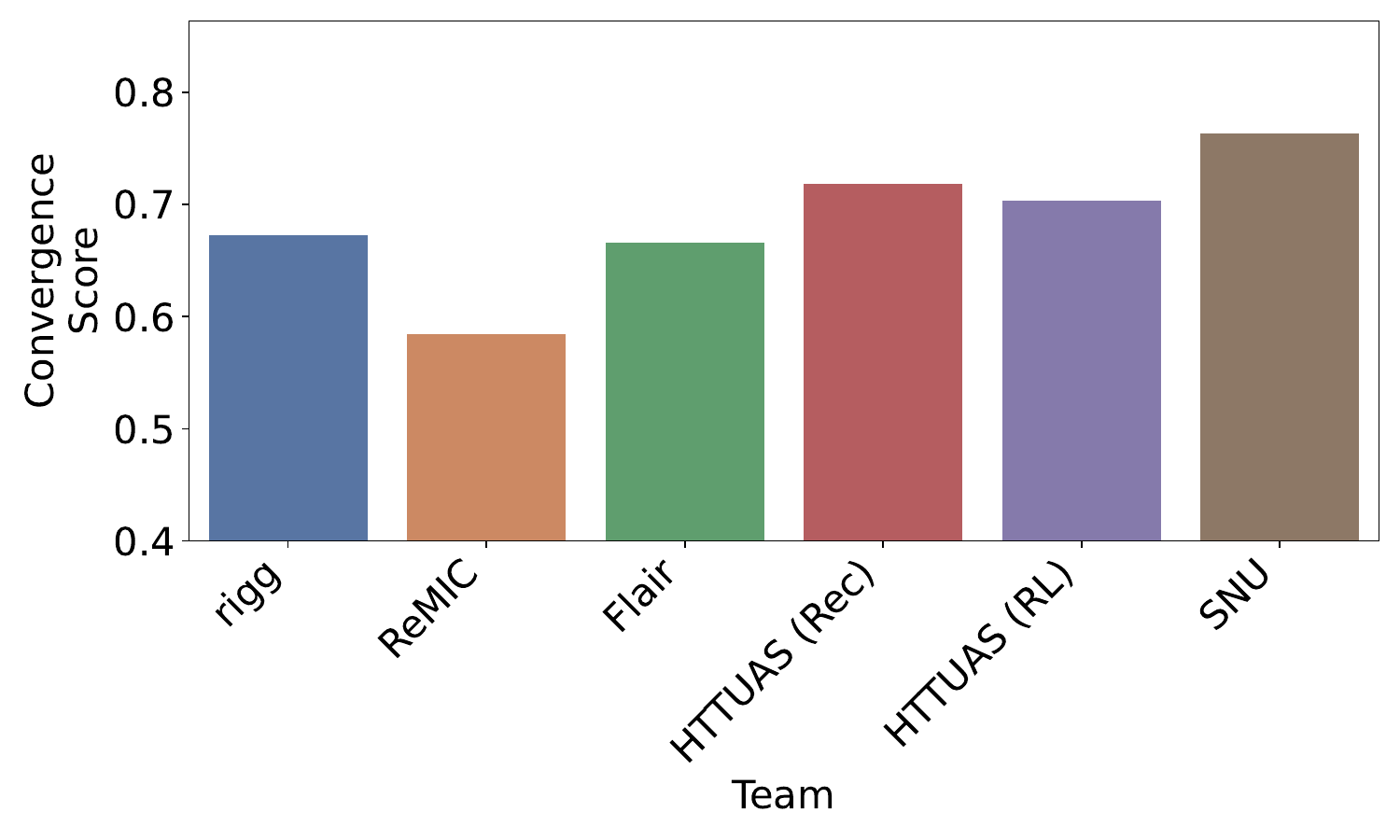}
        \caption{Convergence Score for each team. SNU's approach outperforms all others in this regard.}
        \label{fig:cefficiency}
        \end{figure}

        In terms of communication efficiency, FedPOD (SNU) also has the best convergence score \textbf{by a significant margin} (Figure \ref{fig:cefficiency}). In terms of the other methods, HTTUAS (REC) provides a solid alternative with slightly higher DSC values than HTTUAS (RL) but with more variability in HD95, suggesting a trade-off between segmentation accuracy and robustness. Flair shows good DSC results but has higher variability and below average communication efficiency, reflecting that its dynamic hyper-parameter adjustments and clustering strategy might not be worth the computational overload.

        \begin{table*}[ht!]
        \setlength{\tabcolsep}{10pt}
        \centering
        \caption{Final cumulative rankings of teams in the FeTS Challenge 2024. The table displays two rankings: A. one that includes convergence score---i.e. \textbf{Overall Ranking} ($w = 3$), accounting for computational efficiency. B. one that excludes convergence score---i.e. \textbf{Segmentation-Only} ($w = 0$), evaluating exclusively on segmentation quality.}
       \begin{tabular}{c|c c|c c}
        \hline
        \multirow{2}{*}{\textbf{Team}} & \multicolumn{2}{c|}{\emph{\textbf{Overall Ranking} ($w = 3$)}} & \multicolumn{2}{c}{\emph{\textbf{Segmentation-Only} ($w = 0$)}} \\
        & \textbf{Rank} & \textbf{Score} ($R_t$) & \textbf{Rank} & \textbf{Score} ($R_t$) \\
        \hline
        \textcolor{SNU}{SNU} & \textbf{1} & \textbf{2.317} & 2 & 2.976 \\
        \textcolor{HTTUAS-REC}{HTTUAS (Rec)} & 2 & 2.744 & 3 & 3.116 \\
        \textcolor{rigg}{rigg} & 3 & 3.206 & \textbf{1} & \textbf{2.809} \\
        \textcolor{HTTUAS-RL}{HTTUAS (RL)} & 4 & 3.337 & 4 & 3.505 \\
        \textcolor{Flair}{Flair} & 5 & 4.402 & 5 & 4.102 \\
        \textcolor{ReMIC}{ReMIC} & 6 & 4.994 & 6 & 4.491 \\
        \hline
        \end{tabular}
        \label{tab:split_ranking}
        \end{table*}
        
        The final ranking of each team is determined by summing all individual rankings and computing the cumulative ranking score as described in Equation~\eqref{eq:ranking_formula}. This approach ensures that both segmentation accuracy and computational efficiency are considered in a balanced manner. The winner of FeTS Challenge 2024 is the FedPOD method of SNU, while the FedPID method is also on the top 3~\ref{tab:split_ranking}. Even if we remove the communication efficiency from the ranking assessment, these two methods remain on the top 3 of the leaderboard. As both are based on the PID-controller, this showcases the effectiveness of this foundation for weight aggregation techniques.

    \section{Discussion}
    
        The MICCAI FeTS Challenge 2024 explored FL for glioma sub-region segmentation in brain mpMRI scans, focusing on innovations in weight aggregation. By analyzing the results of six distinct approaches, we identify trends and insights, highlighting both strengths and areas for improvement. 
        
        Two particular methods were of highest interest: FedPID and FedPOD, which achieved first and second ranking respectively if we only account for performance on the DSC and HD95, then third and first if we also account for communication efficiency (Table~\ref{tab:split_ranking}). Their relative variability on performance in the box plots was also the most similar in terms of robustness (Figure ~\ref{fig:combined_dsc_hd}). The similarity in performance may be due to the fact that both these methods build on a predecessor from FeTS Challenge 2022 \citep{zenk2025towards}, FedPIDAvg \citep{machler2022fedpidavg}, whose first iteration was FedCostWAvg \citep{machler2021fedcostwavg} presented in FeTS Challenge 2021. In both these challenges, the respective method achieved competitive ranking, and now two top performers are building on it further, suggesting this method presents a highly reliable baseline for innovation on the algorithm of FL.
    
        Methods such as HTTUAS (Rec) and Flair illustrate the challenge of balancing segmentation accuracy with communication efficiency. HTTUAS (Rec) performed competitively in DSC while maintaining lower variability in HD95. Flair’s adaptive strategies yielded moderate results but were hindered by higher computational demands. Communication efficiency was a critical metric in the final ranking, in which FedPOD outperformed all others by a significant margin, underscoring its relevance for real-world FL applications. In particular, it showcased what could be a critical innovation for FL efficiency: reframing the process as deployable Kubernetes units. This could be a highly valuable modification to the algorithm for multi-site collaborations, especially under resource constraints.

        The observed variability across aggregation strategies can be partly explained in medical imaging context: clinical datasets are inherently heterogeneous across sites not only in size but also in characteristics, reflecting differences in scanners, acquisition protocols, and patient populations, and outliers are not infrequent. On the one hand, HTTUAS methods are based on SimAgg assumes similarity of local models to global average, which given non-IID settings of medical imaging is often an incorrect assumption that does not acknowledge the presence of outliers. SimAgg-based methods thus favor high-performing collaborators and may boost overall segmentation accuracy but risk overlooking smaller institutions, raising questions of fairness. On the other hand, algorithms such as FedPID and FedPOD regulate updates across rounds using controller-inspired dynamics, which stabilizes learning under the non-uniform conditions of medical imaging, and effectively model outliers with a Poisson distribution. While FedPID and its predecessors exclude these outliers, FedPOD has devised a way to include them effectively in the training.
        
        Future work may apply FedPOD and other PID-based methods beyond simulation-based evaluations to real-world deployments, and assess these highlighted benefits in settings of realistic device heterogeneity and communication. FedPOD's inclusion of outliers makes it an interesting candidate for fairness studies, while its compatibility of Kubernetes further enables its application on large-scale real world scenarios. This challenge, as its predecessors, highlights that FL algorithms are ripe for innovation, and sets a foundation for continued improvement.

%%%%%%%%%%%%%%%%%%%%%%%%%%%%%%%%%%%%%%%%%%%%%%%%%%%%%%%%%%%%%%%%%%%%%%%
% Mandatory Sections. Please complete, especially for final publication
%%%%%%%%%%%%%%%%%%%%%%%%%%%%%%%%%%%%%%%%%%%%%%%%%%%%%%%%%%%%%%%%%%%%%%%

% Data availability
\data{The imaging data used for the FeTS Challenge 2024 are derived from the publicly available BraTS 2021 dataset. Access to the dataset in the format used for this challenge can be obtained through The Cancer Imaging Archive: \url{https://www.cancerimagingarchive.net/analysis-result/rsna-asnr-miccai-brats-2021/}.}

% Ethical Standards.
% Please edit with the appropriate ethics considerations for your work. Include any pertinent IRB information, etc.
%
% Please note that the submission requirements included:
% The work presented must follow appropriate ethical standards in conducting research and writing the manuscript, following all applicable laws and regulations regarding treatment of animals or human subjects.
\ethics{This work involves no new human or animal subject data collection. All imaging data used in the FeTS Challenge 2024 are from the publicly available BraTS 2021 dataset, which was de-identified and ethically approved under the respective Institutional Review Boards (IRBs) of contributing institutions. Data use complies with HIPAA and GDPR regulations. No additional IRB approval was necessary for this study. All data derivatives used in the challenge preserve and respect the original dataset’s licensing and ethical constraints.}

\ethics{The FeTS Challenge 2024 adheres to rigorous ethical standards, ensuring compliance with data privacy and consent regulations. The challenge exclusively uses the BraTS 2021 dataset, which is publicly available and open source. Since no new patient data are collected or shared, the challenge does not involve privacy risks or concerns related to data protection. Furthermore, the challenge investigates FL, a paradigm that, when deployed in the real world, allows data to remain within originating institutions, thereby preserving patient privacy and complying with regulations such as HIPAA and GDPR.

All human data included in the challenge originate from the BraTS dataset, where Institutional Review Board (IRB) approval had been previously obtained. Data collection and processing adhered to protocols ensuring subjects' informed consent or opt-in mechanisms, as per institutional policies.

For derivative data used within the challenge, such as preprocessed MRI scans, compliance with the original dataset’s licensing terms is maintained, ensuring ethical use and redistribution of data.}

% Acknowledgements.
% Please include any funding, intellectual contributions not included in the authorship, and any other acknowledgements.
\acks{Research reported in this publication was partially supported by the National Cancer Institute (NCI) of the National Institutes of Health (NIH) under the award numbers U24CA189523 and U01CA242871. Computational resources used in this research were partially supported in part by Lilly Endowment, Inc., through its support for the Indiana University Pervasive Technology Institute. The content of this publication is solely the responsibility of the authors and does not represent the official views of the NIH or any other funding body.}

% Conflict of Interest
% Declaration of possible conflicts of interest: Authors must disclose any financial, organisational, commercial or personal conflicts of interest that might bias their work.
% If no conflicts, please say "We declare we don't have conflicts of interest."
\coi{The Intel-affiliated authors (B. Edwards, P. Foley, M. Sheller, and K. Ta) disclose potential competing interests as employees of Intel Corporation. Intel may develop proprietary software that could be perceived as related to the OpenFL open-source project that is the FL backbone of this work. Additionally, this study highlights the feasibility of federated learning for brain tumor segmentation, a domain where Intel could benefit from increased demand for relevant computational products. The remaining authors declare no competing interests.}

\bibliography{sample}

\end{document}